\documentclass[10pt,twocolumn,letterpaper]{article}

\usepackage[pagenumbers]{cvpr} 

\usepackage{graphicx}
\usepackage{amsmath}
\usepackage{amssymb}
\usepackage{booktabs}

\usepackage{graphicx}
\usepackage{amsmath}
\usepackage{amssymb}
\usepackage{booktabs}
\usepackage{times}
\usepackage{epsfig}
\usepackage{graphicx}
\usepackage{amsmath}
\usepackage{amssymb}
\usepackage{times,enumitem}
\usepackage{epsfig}
\usepackage{graphicx}
\usepackage{amsmath}
\usepackage{amssymb}
\usepackage{mathtools}
\usepackage[table,xcdraw]{xcolor}
\usepackage{url}
\makeatletter
\@namedef{ver@everyshi.sty}{}
\makeatother
\usepackage{tikz}
\usepackage{comment}
\usepackage{amsmath,amssymb} 
\usepackage{color}
\usepackage{graphicx}
\usepackage{amsmath}
\usepackage{amssymb}
\usepackage{booktabs}
\usepackage{cuted}
\usepackage{capt-of}
\usepackage{comment}
\usepackage{breqn}
\usepackage{amsmath}
\usepackage{multirow}
\usepackage{enumitem}
\usepackage{makecell}
\usepackage{enumitem}
\usepackage{xcolor}

\usepackage[pagebackref,breaklinks,colorlinks]{hyperref}
\usepackage[capitalize]{cleveref}
\crefname{section}{Sec.}{Secs.}
\Crefname{section}{Section}{Sections}
\Crefname{table}{Table}{Tables}
\crefname{table}{Tab.}{Tabs.}

\def\cvprPaperID{5141} 
\def\confName{CVPR}
\def\confYear{2023}

\begin{document}



\title{Ambiguous Medical Image Segmentation using Diffusion Models}
\author{Aimon Rahman\textsuperscript{1}~~~~~~Jeya Maria Jose Valanarasu\textsuperscript{1}~~~~~~Ilker Hacihaliloglu$^{2}$~~~~~~Vishal M. Patel$^{1}$ \\
$^1$Johns Hopkins University~~~~~~ $^2$University of British Columbia\\
{\tt\small arahma30@jhu.edu}
}
\maketitle

\begin{abstract}

Collective insights from a group of experts have always proven to outperform an individual's best diagnostic for clinical tasks. For the task of medical image segmentation, existing research on AI-based alternatives focuses more on developing models that can imitate the best individual rather than harnessing the power of expert groups. In this paper, we introduce a single diffusion model-based approach that produces multiple plausible outputs by learning a distribution over group insights. 
Our proposed model generates a distribution of segmentation masks by leveraging the inherent stochastic sampling process of diffusion using only minimal additional learning. We demonstrate on three different medical image modalities- CT, ultrasound, and MRI that our model is capable of producing several possible variants while capturing the frequencies of their occurrences. 
Comprehensive results show that our proposed approach outperforms existing state-of-the-art ambiguous segmentation networks in terms of accuracy while preserving naturally occurring variation. We also propose a new metric to evaluate the diversity as well as the accuracy of segmentation predictions that aligns with the interest of clinical practice of collective insights. Implementation code: \href{https://github.com/aimansnigdha/Ambiguous-Medical-Image-Segmentation-using-Diffusion-Models}{https://github.com/aimansnigdha/Ambiguous-Medical-Image-Segmentation-using-Diffusion-Models}.
\vskip -18 pt
\end{abstract}
\section{Introduction}
\vskip -5 pt

Diagnosis is the central part of medicine, which heavily relies on the individual practitioner assessment strategy.  Recent studies suggest that misdiagnosis with potential mortality and morbidity is widespread for even the most common health conditions \cite{sonderegger2000diagnostic, kurvers2016boosting}. Hence, reducing the frequency of misdiagnosis is a crucial step towards improving healthcare. Medical image segmentation, which is a central part of diagnosis, plays a crucial role in clinical outcomes. Deep learning-based networks for segmentation are now getting traction for assisting in clinical settings, however, most of the leading segmentation networks in the literature are deterministic \cite{ronneberger2015u,milletari2016v,jha2019resunet++,gu2019net, li2020transformation,rahman2022orientation,rahman2022simultaneous}, meaning they predict a single segmentation mask for each input image. Unlike natural images, ground truths are not deterministic in medical images as different diagnosticians can have different opinions on the type and extent of an anomaly \cite{alpert2004quality,doi1999computer,qiu2020modal, monteiro2020stochastic}. Due to this, the diagnosis from medical images is quite challenging and often results in a low inter-rater agreement \cite{jensen2019improving,ji2021learning,visser2019inter}. Depending on only pixel-wise probabilities and ignoring co-variances between the pixels might lead to misdiagnosis. In clinical practice, aggregating interpretations of multiple experts have shown to improve diagnosis and generate fewer false negatives \cite{wolf2015collective}.

\begin{figure}[t!]
\centering
\includegraphics[width=\linewidth,scale=1]
{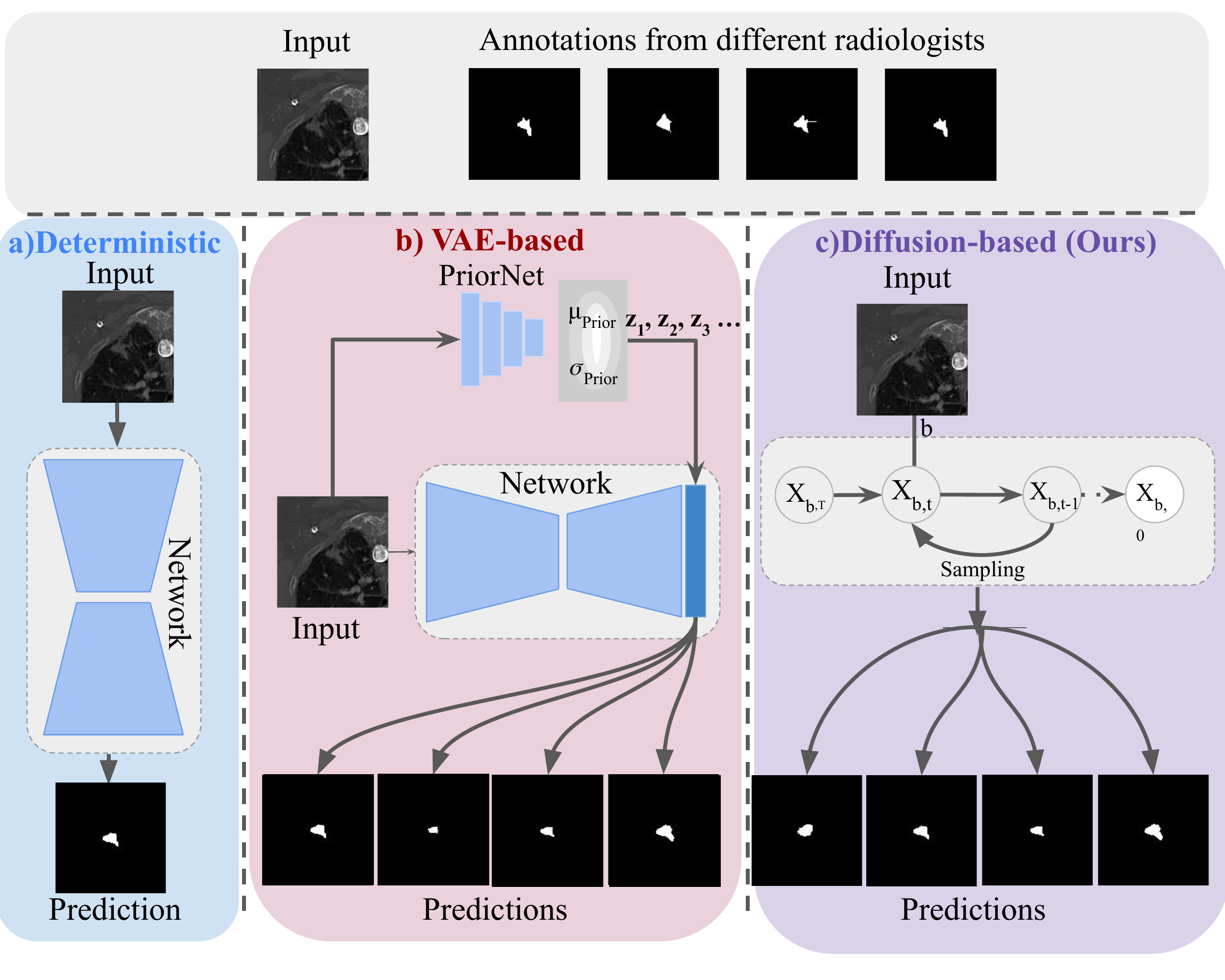}
\vspace{-2em}
\caption{ a) Deterministic networks produce a single output for an input image. b) c-VAE-based methods encode prior information about the input image in a separate network and sample latent variables from there and inject it into the deterministic segmentation network to produce stochastic segmentation masks. c)  In our method the diffusion model learns the latent structure of the segmentation as well as the ambiguity of the dataset by modeling the way input images are diffused through the latent space. Hence our  method does not need an additional prior encoder to provide latent variables for multiple plausible annotations.}
\label{Fig:intro}
\vskip -21pt
\end{figure}

In fact, utilizing the aptitude of multiple medical experts has been a part of long-standing clinical traditions such as case conferences, specialist consultations, and tumor boards. By harnessing the power of collective intelligence, team-based decision-making provides safer healthcare through improved diagnosis \cite{radcliffe2019collective,kurvers2016boosting}. Although collective insight is gaining traction in healthcare for its potential in enhancing diagnostic accuracy, the method and its implication remain poorly characterized in automated medical vision literature.  It has been suggested that the use of artificial intelligence can optimize these processes while considering physician workflows in clinical settings \cite{radcliffe2019collective}.

In recent times, there has been an outstanding improvement in specialized deterministic models for different medical image segmentation tasks \cite{ronneberger2015u,valanarasu2022unext,zhou2018unet++,valanarasu2021kiu, chen2021transunet,  valanarasu2021medical, valanarasu2020kiu}. Deterministic models are notorious for choosing the most likely hypothesis even if there is uncertainty which might lead to sub-optimal segmentation. To overcome this, some models incorporate pixel-wise uncertainty for segmentation tasks, however, they produce inconsistent outputs \cite{kendall2015bayesian,kendall2017uncertainties}. Conditional variational autoencoders (c-VAE) \cite{sohn2015learning}, a conditional generative model, can be fused with deterministic segmentation networks to produce unlimited numbers of predictions by sampling from the latent space conditioned on the input image. Probabilistic U-net and its variants use this technique during the inference process. Here, the latent spaces are sampled from a prior network which has been trained to be similar to c-VAE \cite{kohl2018probabilistic,kohl2019hierarchical,baumgartner2019phiseg}. This dependency on a prior network as well as injecting stochasticity only at the highest resolution of the segmentation network produces less diverse and blurry segmentation predictions \cite{selvan2020uncertainty}. To overcome this problem, we introduce a single inherently probabilistic model without any additional prior network that represents the collective intelligence of several experts to leverage multiple plausible hypotheses in the diagnosis pipeline (visualized in Figure \ref{Fig:intro}).   

Diffusion probabilistic models are a class of generative models consisting of Markov chains trained using variational inference \cite{ho2020denoising}. The model learns the latent structure of the dataset by modeling the diffusion process through latent space. A neural network is trained to denoise noisy image blurred using Gaussian noise by learning the reverse diffusion process  \cite{song2020improved}. Recently, diffusion models have been found to be widely successful for various tasks such as image generation \cite{choi2021ilvr}, and inpainting \cite{lugmayr2022repaint}. Certain approaches have also been proposed to perform semantic segmentation using diffusion models \cite{baranchuk2021label,wolleb2021diffusion}. Here, the stochastic element in each sampling step of the diffusion model using the same pre-trained model paves the way for generating multiple segmentation masks from a single input image. However, there is still no exploration of using diffusion models for ambiguous medical image segmentation despite its high potential. In this paper, we propose the CIMD (\textbf{C}ollectivly \textbf{I}ntelligent \textbf{M}edical \textbf{D}iffusion), which addresses ambiguous segmentation tasks of medical imaging. First, we introduce a novel diffusion-based probabilistic framework that can generate multiple realistic segmentation masks from a single input image. This is motivated by our argument that the stochastic sampling process of the diffusion model can be harnessed to sample multiple plausible annotations. The stochastic sampling process also eliminates the need for a separate ‘prior’ distribution during the inference stage, which is critical for c-VAE-based segmentation models to sample the latent distribution for ambiguous segmentation. The hierarchical structure of our model also makes it possible to control the diversity at each time step hence making the segmentation masks more realistic as well as heterogeneous. Lastly, in order to assess ambiguous medical image segmentation models, one of the most commonly used metrics is known as GED (Generalized Energy Distance), which matches the ground truth distribution with prediction distribution. In real-world scenarios for ambiguous medical image segmentation, ground truth distributions are characterized by only a set of samples. In practice, the GED metric has been shown to reward sample diversity regardless of the generated samples' fidelity or their match with ground truths, which can be potentially harmful in clinical applications \cite{kohl2019hierarchical}. In medical practice, individual assessments are manually combined into a single diagnosis and evaluated in terms of sensitivity. When real-time group assessment occurs, the participant generates a consensus among themselves. Lastly, the minimum agreement and maximum agreement among radiologists are also considered in clinical settings. Inspired by the current practice in collective insight medicine, we coin a new metric, namely the CI score (\textbf{C}ollective \textbf{I}nsight) that considers total sensitivity, general consensus, and variation among radiologists. In summary, the following are the major contributions of this work:

\begin{itemize}[noitemsep]

\vspace{-.8em}
    \item We propose a novel diffusion-based framework: Collectively Intelligent Medical Diffusion (CIMD), that realistically models heterogeneity of the segmentation masks without requiring any additional network to provide prior information during inference unlike previous ambiguous segmentation works.

    \item  We revisit and analyze the inherent problem of the current evaluation metric, GED for ambiguous models and explain why this metric is insufficient to capture the performance of the ambiguous models. We introduce a new metric inspired by collective intelligence medicine, coined as the CI Score (\textbf{C}ollective \textbf{I}nsight).

    \item We demonstrate across three medical imaging modalities that CIMD performs on par or better than the existing ambiguous image segmentation networks in terms of quantitative standards while producing superior qualitative results.
\vskip -20 pt
\end{itemize}
\vspace{-1em}
\section{Related Work}
\vspace{-.3em}
\noindent \textbf{Ambiguous Image Segmentation.} Previous work \cite{kendall2015bayesian} models the ambiguity using approximate Bayesian inference over the network weights. However, the method is shown to produce samples that only vary pixel by pixel and can not capture the complex correlation structure of the ground truth distribution \cite{kohl2018probabilistic}. Probabilistic U-net is capable of capturing distribution over multiple annotations that can produce a wide variety of segmentation maps from a single image \cite{kohl2018probabilistic}. The model is a combination of a U-net with a conditional variational auto-encoder that uses its stochasticity to produce an unlimited number of plausible hypotheses. This method has been shown to produce samples with limited diversity as the stochasticity is only injected in the highest resolution of the backbone segmentation network, hence the network chooses to ignore the random draws from the latent space \cite{baumgartner2019phiseg}. 
To increase sample diversity, PHi-SegNet \cite{baumgartner2019phiseg} and Hierarchical Probabilistic  U-Net \cite{kohl2019hierarchical} incorporate a series of hierarchical latent spaces to sample the feature maps. The key element of their works is that the backbone networks rely on variational inference to produce multiple annotations for an image by sampling from a distribution, which, if not sufficiently complex, may not produce realistic samples \cite{bhat2022generalized}. The diversity of segmentation masks of these c-VAE-like models relies on an axis-aligned Gaussian latent posterior distribution, which can be too restrictive and not expressive enough to model the rich variations \cite{selvan2020uncertainty}.

\noindent \textbf{Diffusion Model for Image Segmentation.} Diffusion models have recently shown remarkable potential in various segmentation tasks \cite{amit2021segdiff,brempong2022denoising,baranchuk2021label,bandara2022ddpm} including medical images \cite{wolleb2021diffusion,wu2022medsegdiff,kim2022diffusion,guo2022accelerating}. In fact, the stochastic sampling process of the diffusion model has been utilized to generate an implicit ensemble of segmentations that ultimately boosts the segmentation performance \cite{wolleb2021diffusion}. However, the model is only trained using a single segmentation mask per input image, hence the model has no control over the variation as well as the produced masks are not necessarily diverse. To the best of our knowledge, CIMD is the first network specifically designed to model the ambiguity of medical images by harnessing its random sampling process. Moreover, the diffusion model’s hierarchical structure makes it possible to govern the ambiguity at each time step, thereby eliminating the problem of low diversity of previous methods.
 \vspace{-1em}
\section{Proposed Method}
\subsection{Diffusion Model}
Diffusion probabilistic models have gained a lot of attention in recent years due to their ability to generate extraordinarily high-quality images compared to Generative Adversarial Networks (GANs), Variational Autoencoders (VAEs), autoregressive models, and flows. Due to their superior ability in extracting key semantics from training images, diffusion models are also being utilized for image segmentation \cite{wolleb2021diffusion,amit2021segdiff}. Additionally, using random Gaussian noise, it is possible to generate a distribution of segmentation rather than a deterministic output. In this section, we present a brief overview of the diffusion model framework.\\ 

\begin{figure*}[htbp]
\centering
\includegraphics[width=.8\linewidth]
{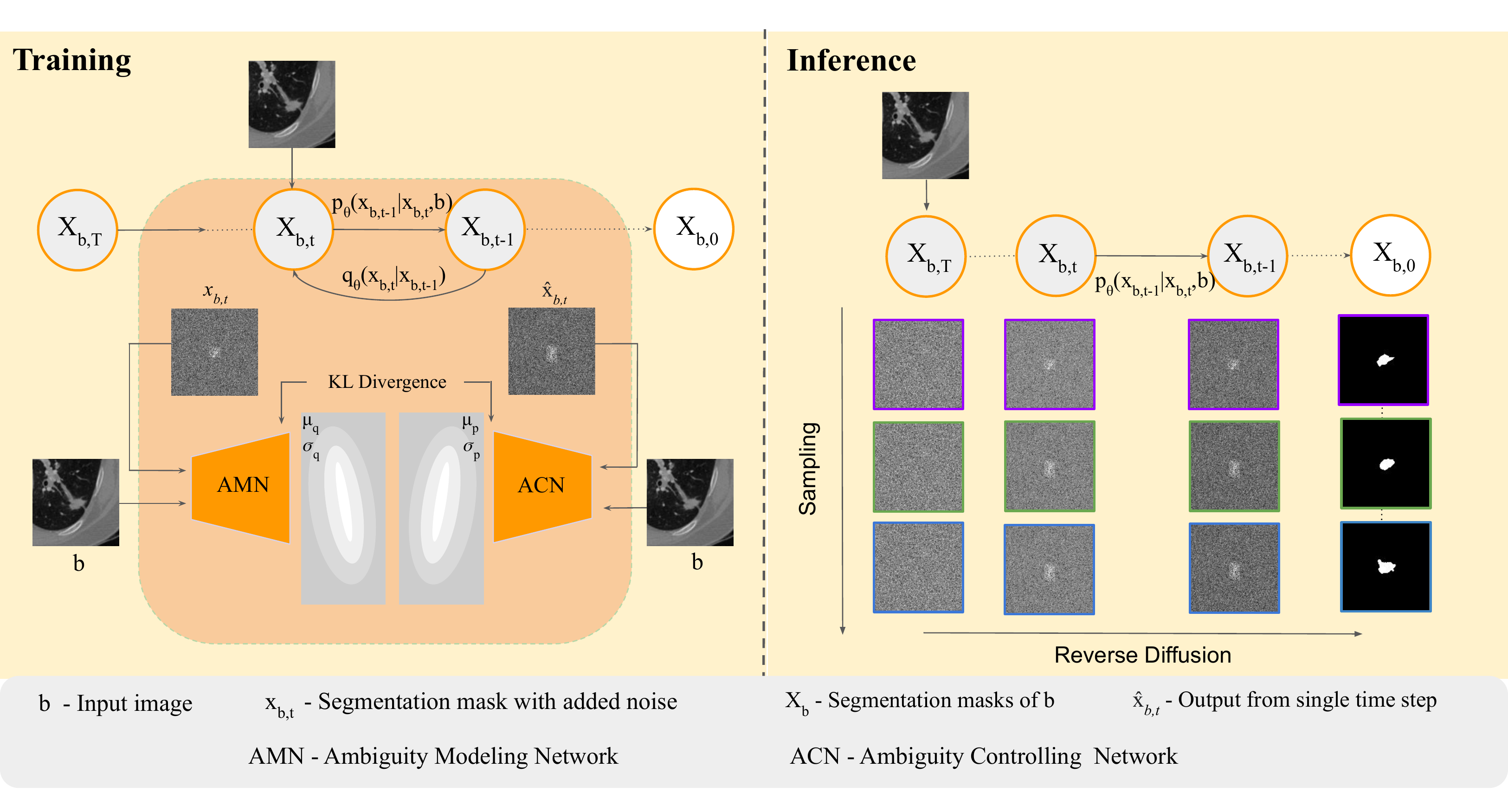}
\vskip-15pt
\caption{
The graphical model of a) sampling and b) training procedure of our method. In the training phase, for every step $t$, the anatomical structure is induced by adding the input image $b$ to the noisy segmentation mask $x_{b,t}$. Sampling $n$ times with different Gaussian noise, $n$ different plausible masks are generated.
}
\vspace{-1em}
\label{Fig:details}
\end{figure*}

\noindent \textbf{Gaussian Diffusion Process.} Diffusion models perform variational inference on a Markovian process using $T$ timesteps to learn the training data distribution $p(x_0)$. The framework consists of a forward and a reverse process. During the forward process for each timestep in $T$, Gaussian noise is added to the image $x_0 \sim p(x_0) $ until the image becomes an isotropic Gaussian. This forward noising process is denoted as,
\setlength{\belowdisplayskip}{0pt} \setlength{\belowdisplayshortskip}{0pt}
\setlength{\abovedisplayskip}{0pt} \setlength{\abovedisplayshortskip}{0pt}
\begin{equation}
\label{eqn:one}
 q(x_t|x_{t-1}) = \mathcal{N}(x_t;\sqrt{\alpha_t}x_{t-1},(1 - \alpha_t)I),
\end{equation}
where  $(x_0, x_1,.., x_T )$ denotes $T$ steps in the Markov chain and $\alpha$ is the noise scheduler that controls the variance of the noise. In the reverse process, a neural network ($f_\theta$) is used to create a sequence of incremental denoising operations to obtain back the clean image. $f_\theta$ learns the parameters of the reverse distribution $p(x_{t-1}|x_t) :=  \mathcal{N}(x_{t-1}: \mu_\theta(x_t,t), \sum_\theta(x_t, t))$. The parameters for $f_\theta$ are obtained by minimizing the KL-divergence between the forward and the reverse distribution for all timesteps. The optimization requires sampling from the distribution $q(x_t|x_{t-1})$ that subsequently requires the knowledge of $x_{t-1}$. Given $x_0$, the marginal distribution of $x_t$ can be obtained by marginalizing out the intermediate latent variable as,
\begin{equation}
\label{eqn:two}
    q(x_t|x_0) = \mathcal{N}(x_t;\sqrt{\gamma_t}x_0,(1 - \gamma_t)I).
\end{equation}
Here, $\gamma_t = \Pi_{i=1}^{t} \alpha_i$. However, minimizing the KL-divergence between the forward and the reverse distribution can be further simplified by using a posterior distribution $q(x_t|x_{t-1}, x_0)$ instead \cite{sohl2015deep}. The posterior distribution can be derived using  Eq. \ref{eqn:one} and \ref{eqn:two} under the Markovian assumptions,
\begin{equation}
    q(x_{t-1}|x_t, x_0) = \mathcal{N} (x_{t-1}, \mu, \sigma^2 I),
\end{equation}
where, $\mu(x_t, x_0)=\frac{\sqrt{\gamma_{t-1}}(1-\alpha_t)}{1-\gamma_t} x_0+\frac{\sqrt{\alpha_t} (1-\gamma_{t-1})}{1-\gamma_t} x_t$ and $\sigma^2 = \frac{(1 - \gamma_{t-1})(1-\alpha_t)}{1-\gamma_t}$.
This posterior distribution is then utilized during the parameterization of the reverse Markov chain for formulating a variational lower bound on the log-likelihood of the reverse chain. During optimization, the covariance matrix for both distributions $q(x_{t-1}|x_t, x_0)$ and $p(x_{t-1}|x_t)$ are considered the same, and the mean of the distributions is predicted by $f_\theta$. The denoising model $f_\theta$ takes noisy image $x_t$ as input which is denoted by,
\begin{equation}
    x_t = \sqrt{\gamma}x_0+ \sqrt{1-\gamma}\epsilon,
\end{equation}
where, $\epsilon = \mathcal{N}(0,I)$. Now, the combination of $p$ and $q$ is a variational auto-encoder \cite{kingma2013auto} and the variational lower bound ($V_{lb}$) can be expressed as,
\begin{equation}
\label{eqn:vlb}
    \mathcal{L}_{vlb} := \mathcal{L}_0 + \mathcal{L}_1 + ... + \mathcal{L}_{T-1} + \mathcal{L}_{T}
\end{equation}
\begin{equation}
    \mathcal{L}_0 := - \log p_\theta(x_0|x_1) 
\end{equation}
\begin{equation}
    \mathcal{L}_{t-1} := D_{KL}(q(x_{t-1}|x_t, x_0) \| p_{\theta}(x_{t-1}|x_t))
\end{equation}
\begin{equation}
  \mathcal{L}_{T} := D_{KL}(q(x_T |x_0) \| p(x_T )).  
\end{equation}
However, the training objective can be further simplified as \cite{ho2020denoising},
\begin{equation}
\label{eqn:simp}
   \mathcal{L}_{simple} =  \mathbb{E}_{x_0,\epsilon} |f_\theta(\Tilde{x},t)-\epsilon|_2^2,
\end{equation}
where $\epsilon=\mathcal{N}(0,I)$. Now, log-likelihood is considered a good metric to evaluate generative models, and optimizing log-likelihood has been proven to force the model to capture all the data distribution \cite{razavi2019generating} as well as improve sample quality \cite{henighan2020scaling}. Hence, we get the hybrid loss \cite{nichol2021improved} by combining Eq. \ref{eqn:vlb} and \ref{eqn:simp}, 
\begin{equation}
\label{eqn:mod1}
    \mathcal{L}_{hybrid} = \mathcal{L}_{simple} + \lambda \mathcal{L}_{vlb},
\end{equation}
where, $\lambda$ is a regularization parameter, which is used to prevent $\mathcal{L}_{vlb}$ from overwhelming $\mathcal{L}_{simple}$. Inference starts from a Gaussian noise $x_t$ which at each timestep is iteratively denoised to get back $x_{t-1}$ as follows,
\begin{equation}
\label{eqn:mod}
    x_{t-1} \leftarrow \frac{1}{\sqrt{\alpha_t}} \left(x_t - \frac{1- \alpha_t}{\sqrt{1-\gamma_t}}f_\theta(x_t,t)\right)+\gamma_t z,
\end{equation}
where, $z = \mathcal{N}(0,I)$ and $t=T,..,1$.

\begin{figure*}[htbp]

\centering
\includegraphics[width=\linewidth]
{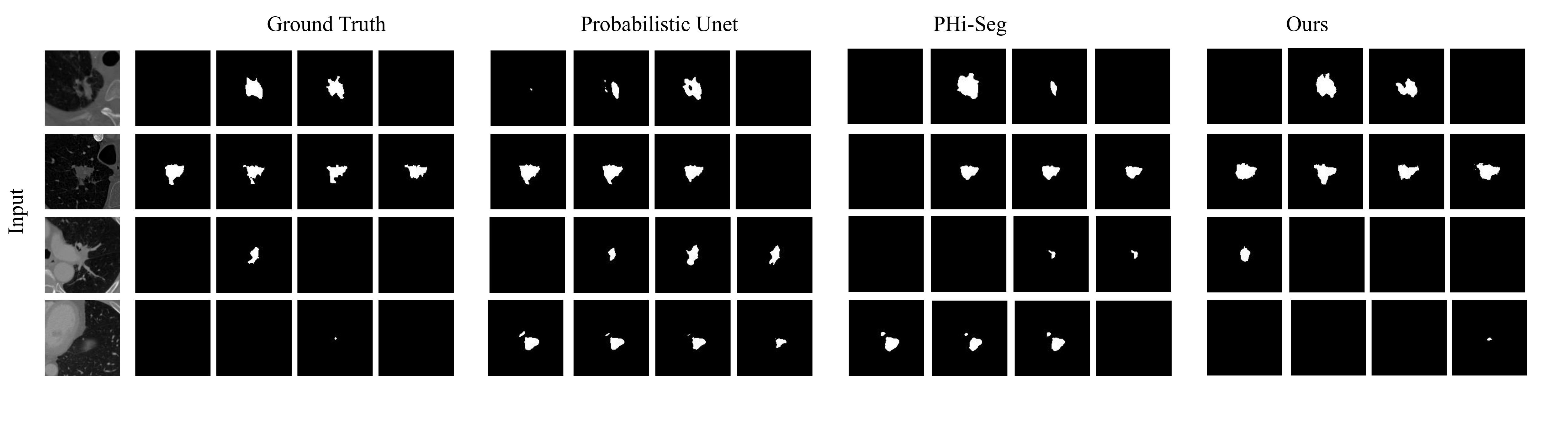}
\vspace{-3.5em}
 \caption{
Comparative qualitative analysis with the two baseline methods -- Probabilistic U-net \cite{kohl2018probabilistic} and PHi-Seg \cite{baumgartner2019phiseg}. Sample images from the LIDC-IDRI dataset with 4 available expert gradings are shown on the left. Note that empty segmentation masks are also valid grading. For a fair comparison, we visualize only the first 4 sampled segmentation masks from the segmentation networks.
}
\vspace{-1em}
\label{Fig:lidc}
\end{figure*}

\subsection{Collectivly Intelligent Medical Diffusion}
Let $b$ be the given image with dimension of $C \times H \times W$ and $x_b$ the corresponding segmentation mask. In a classical diffusion model, the input image $x_b$ is required for training, which would result in an arbitrary segmentation mask $x_0$ when sampled from noise during inference. In contrast to that, to produce a segmentation mask $x_{b,0}$ for a given image $b$, an additional channel is concatenated to the input. This induces the anatomical information by concatenating it as an image prior to $x_b$ and thus defining $X:=b \oplus x_b$. During the noising process $q$, noise is added to the ground truth segmentation $x_b$ only. As the sampling process is stochastic, the diffusion model produces different segmentation masks $x_{b,0}$ for an image $b$. When the diffusion model is trained using only one segmentation mask per input image, the model can implicitly generate an ensemble of segmentation masks that can be used to boost the performance of the model \cite{wolleb2021diffusion}. Now, we model the ambiguity of the ground truths using \textbf{Ambiguity Modelling Network (AMN)}. AMN models the distribution of ground truth masks given an input image. We embed this ambiguity of the segmentation masks in the latent space by parameterizing the weight of AMN by $\nu$, given the image $b$ and the ground truth $x_b$. This probability distribution denoted as $Q$ is modeled as a Gaussian with mean $\mu(b,x_b;\nu) \in R^N$ and variance $\sigma (b,x_b;\nu) \in R^{N \times N}$ where $N$ denotes the low dimensional latent space. The latent space is characterized by,
\begin{equation}
z_q \sim Q(.|b,x_b) = \mathcal{N} (\mu(b,x_b;\nu),\sigma(b,x_b;\nu)). 
\end{equation}
Similarly, we model the ambiguity of predicted masks using \textbf{Ambiguity Controlling Network (ACN)}. ACN models the noisy output from the diffusion model conditioning on an input image. For each time step $t$, assuming $\hat{x_{b}} = f_\theta(\Tilde{x_b},t)$, we estimate the ambiguity of our diffusion model by parameterizing the weight of ACN, $\omega$ as a probability distribution $P$ with mean $\mu(b,\hat{x_b};\omega) \in R^N$ and variance $\sigma (b,\hat{x_b};\omega) \in R^{N\times N}$ as follows
\begin{equation}
z_p \sim P(.|b,\hat{x_{b}}) = \mathcal{N} (\mu(b,\hat{x_{b}},t);\omega),\sigma(b,\hat{x_{b}},t);\omega)). 
\end{equation}
Both networks \textbf{AMN} and \textbf{ACN} are modeled using an axis-aligned gaussian distribution with diagonal covariance matrices. The architectural details of both networks can be found in the supplementary. We penalize the difference between two distributions by imposing a Kullback-Leibler divergence,
\begin{equation}
    \mathcal{L}_{amb} = D_{KL}(Q(z|x_b, b)\|P(z|\hat{x_b},b)).
\end{equation}
Finally, by modifying Eq. \ref{eqn:mod1}, all losses are combined as a weighted sum with a regularizing factor $\beta$ as 
\begin{equation}
    \mathcal{L}_{total} = \mathcal{L}_{simple} + \lambda \mathcal{L}_{vlb} + \beta \mathcal{L}_{amb}. 
\end{equation}
During the sampling process, for $X_t:=b \oplus x_{b,t}$, Eq. \ref{eqn:mod} is modified as,
\begin{equation}
    x_{b,t-1} \leftarrow \frac{1}{\sqrt{\alpha_t}} \left(x_{b,t} - \frac{1- \alpha_t}{\sqrt{1-\gamma_t}}f_\theta(X_t,t)\right)+\gamma_t z,
\end{equation}
where, $z = \mathcal{N}(0,I)$ and $t=T,..,1$.
The graphical model of proposed approach is illustrated in Figure \ref{Fig:details}.
\begin{figure}[htbp]
\centering
\vspace{-1em}
\includegraphics[width=0.75\linewidth]
{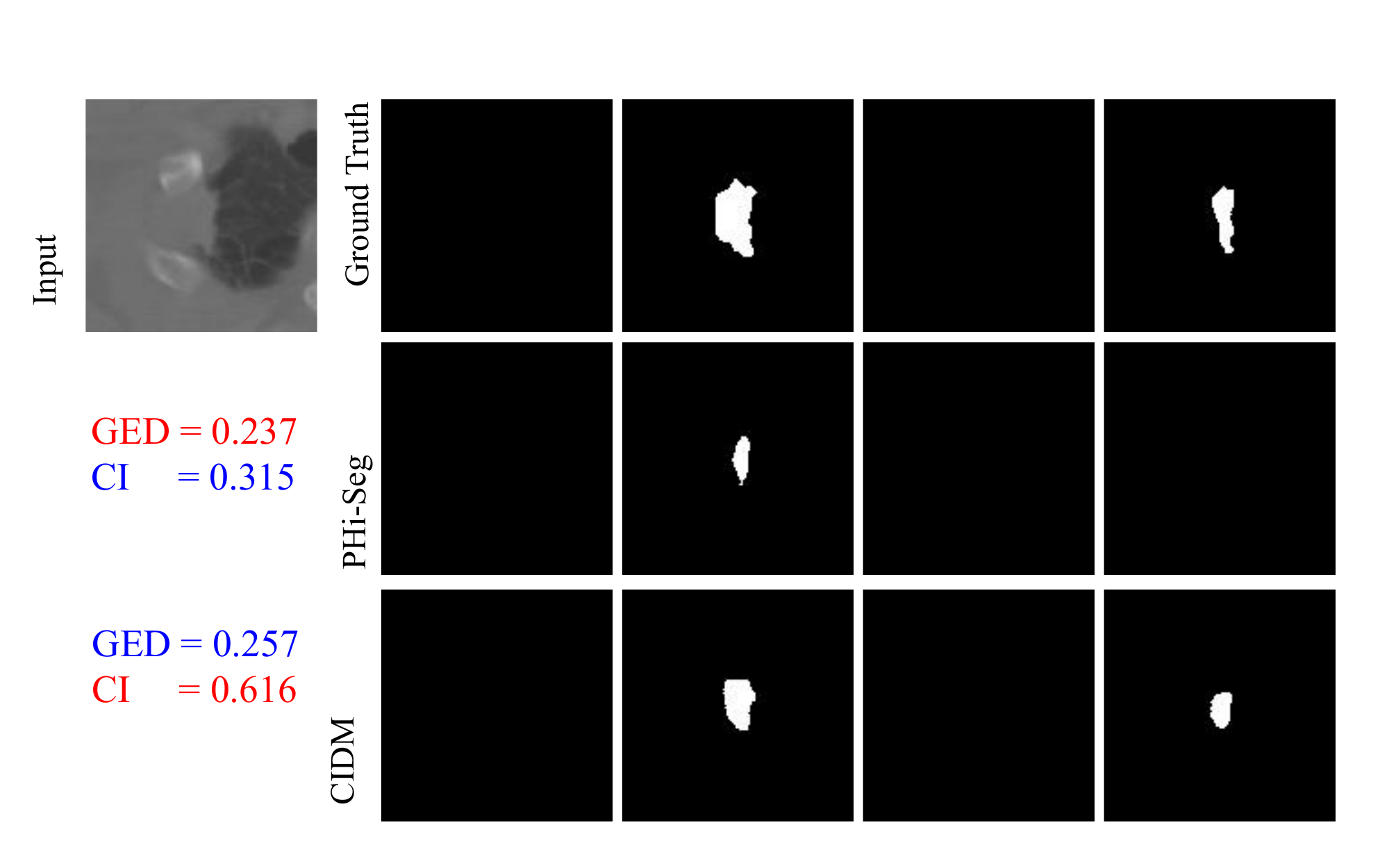}
\vspace{-1em}
 \caption{Visual analysis of the GED vs. the CI score for the LIDC-IDRI lung CT dataset. It can be observed that GED is lower for PHi-Seg even though it failed to segment most of the lesions. However, the combined sensitivity penalizes under segmentation hence the CI score is lower in that case. \textcolor{red}{
Red} corresponds to better and \textcolor{blue}{blue} corresponds to a lower score.}
\label{Fig:gedvci}
\vskip -19pt
\end{figure}

\subsection{Collective Insight Score}
\begin{figure}[htbp]
\centering
\includegraphics[width=1\linewidth]
{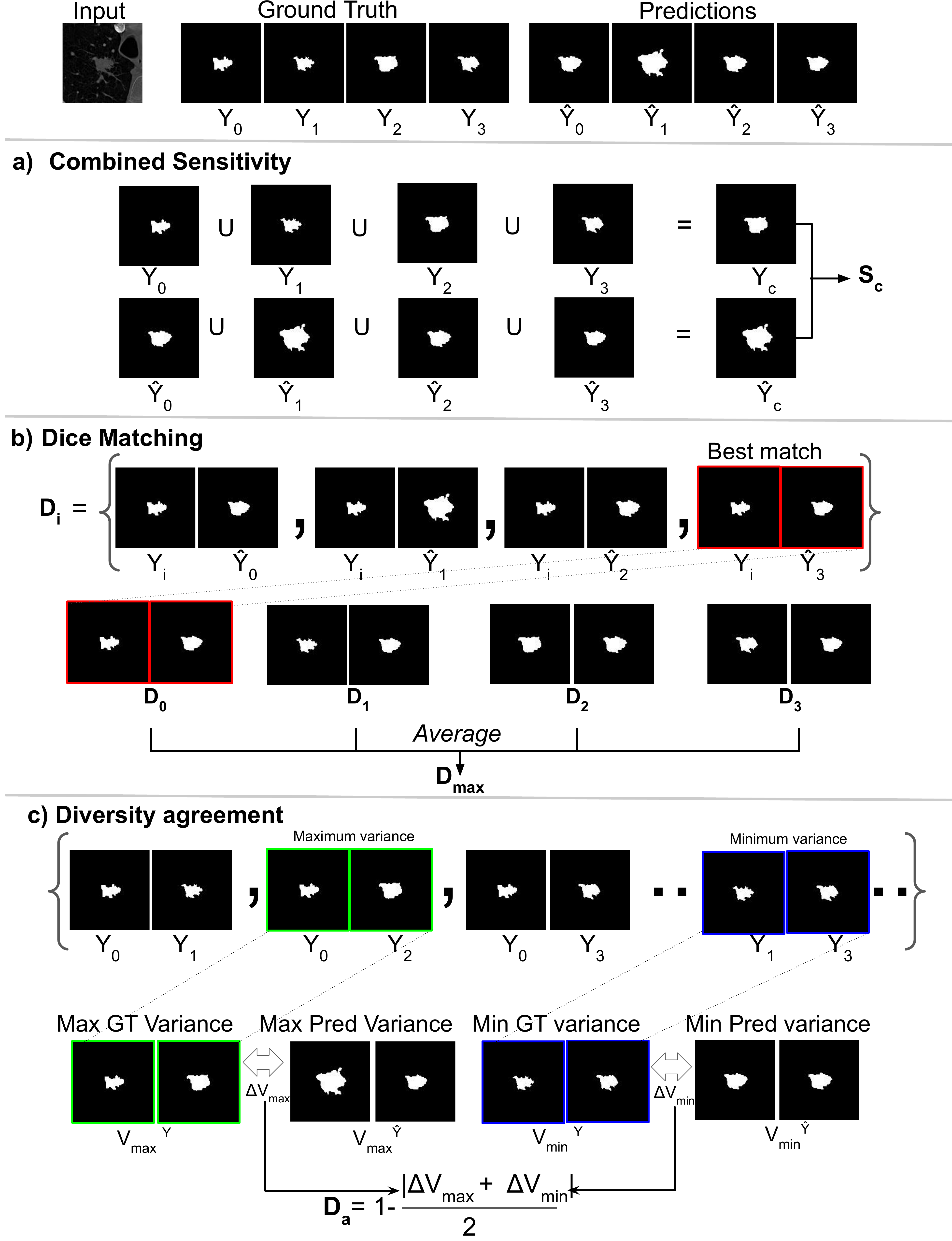}
\vskip-10pt
 \caption{Visual representation of each component a) Combined Sensitivity, b) Dice Matching, and c) Diversity agreement of the proposed CI score metric. GT denotes ground truth and Pred denotes predictions. Here, the number of ground truth $M = 4$. In c) we only demonstrate the variance in ground truth distribution, however, in practice we calculate it for both ground truth and prediction distribution.}
 \vspace{-1em}
\label{Fig:met}
\end{figure}

\vspace{-.3em}
\noindent To recap, ambiguous segmentation models generate a distribution of predictions rather than a deterministic one and are evaluated against a distribution of ground truths. Although Generalized Energy Distance (GED) has been used before for assessing ambiguous segmentation models, this metric has been found to be inadequate as it disproportionately rewards sample diversity regardless of its match with the ground truth samples \cite{kohl2019hierarchical}. This can be potentially dangerous, particularly in pathological cases. In Figure \ref{Fig:gedvci}, we can observe how GED is unduly rewarding PHi-Seg even though CIDM outputs are qualitatively better. To this end, we propose an alternative evaluation metric called the CI score (\textbf{C}ollective \textbf{I}nsight), and explain the motivation behind each component of the metric in this section. The CI score is defined as, 
\begin{equation}
    CI = \frac{3 \times S_c \times D_{max} \times D_a}{S_c + D_{max} + D_a},
\end{equation}
where, $S_c$ denotes the combined sensitivity, $D_{max}$ denotes the maximum Dice matching score and $D_a$ is the diversity agreement score. CI takes the harmonic mean of each component to equalize the weights of each part. In the following sections, true positive, false negative, and false positive are denoted as $TP$, $FN$, and $FP$, respectively.  The collection of all ground truths is denoted as  $\textbf{Y} = \{Y_1,Y_2,..,Y_M \}$ where $Y_1,Y_2,..,Y_M$ corresponds to the individual ground truth of each sample. $M$ here is the number of ground truths. Similarly, the collection of all predictions is denoted as  $\hat{\textbf{Y}} = \{\hat{Y}_1,\hat{Y}_2,..,\hat{Y}_N\}$ where $\hat{Y}_1,\hat{Y}_2,..,\hat{Y}_N$ corresponds to the individual predictions of each sample. $N$ here is the number of predictions. Although the number of ground truths is limited, the model can generate an unlimited number of predictions, hence $M$ and $N$ are not necessarily equal. A visual diagram in Figure \ref{Fig:met} illustrates the operation of each component.
\vskip -5 pt
\noindent \textbf{Combined Sensitivity.} In clinical practice, all the diagnoses from different raters are combined into a single collective decision. The final decision is then usually assessed in terms of true positive rate (sensitivity) \cite{fihn2019collective}. In many branches of medical diagnosis, the primary goal is to maximize the true positive rate while maintaining a tolerable degree of false positive rate \cite{argenziano1998epiluminescence,kurvers2015detection, wolf2013accurate}. Hence, we argue that assessing the combined sensitivity directly aligns with the interest of common clinical practice. Since empty ground truth is also a valid prediction, we consider sensitivity to be 1 in those instances. First, we define the combined ground truth $Y_c$ which is the union of all ground truths maps. Similarly, we define combined predictions $\hat{Y}_c$ which is the union of all prediction maps. $Y_c$ and $\hat{Y}_c$ are mathematically formulated as follows:
\begin{equation}
Y_c = \bigcup_{i=1}^{M} Y_i, \;\;\;\;\;  \hat{Y_c} = \bigcup_{j=1}^{N} \hat{Y_j}.
\end{equation}
We calculate the combined sensitivity $S_c$ between the combined predictions and combined ground truths as follows:
\begin{equation}
S_c(\hat{Y_c},Y_c) = \begin{cases}
        \frac{TP}{TP+FN}, & \text{if } \hat{Y_c} \cup Y_c  \neq \emptyset\\
        1, & \text{if } \text{otherwise}.
    \end{cases}
\end{equation}

\vskip -5 pt

\noindent \textbf{Maximum Dice Matching.}   In medical diagnosis cases, empty sets, which indicate no abnormalities are also valid diagnoses. However, in this case, the Dice metric will be undefined, hence we set Dice = 1 in those cases. Thus, the Dice score is defined as:
\begin{equation}
Dice(\hat{Y},Y) =  \begin{cases}
        \frac{2 |Y \cap \hat{Y}|}{|Y|+|\hat{Y}|}, &\text{if }  Y \cup \hat{Y} \neq \emptyset\\
        1, & \text{ } \text{otherwise}.
    \end{cases}
\end{equation}

In collective intelligence practice, it is common to assess the diagnosis of students against the experts \cite{kunina2015assessing,barnett2019comparative}. We emulate the process by calculating the Dice scores of individual predictions with all the ground truths and then calculating the maximum Dice score among all these pairs. First, we define the set of all  Dice scores $\textbf{D}_i$ for each individual ground truth $Y_i$ as follows:
\vspace{-.2em}
\begin{equation}
    \textbf{D}_i = \{Dice(\hat{Y}_1,Y_i),Dice(\hat{Y}_2,Y_i),...Dice(\hat{Y}_N,Y_i)\},
\end{equation}
where $\textbf{D}_i$ is a collection of Dice scores calculated between each ground truth $Y_i$ and all the provided predictions. Then, we take the maximum Dice score among this set and report the average as the maximum Dice match $D_{max}$. $D_{max}$ is formulated as follows:
\vspace{-1em}
\begin{equation}
    D_{max} = \frac{1}{M} \sum_{i=1}^{M} \max (\textbf{D}_i  ).
\end{equation}

\begin{figure*}[htbp]
\centering
\includegraphics[width=\linewidth]
{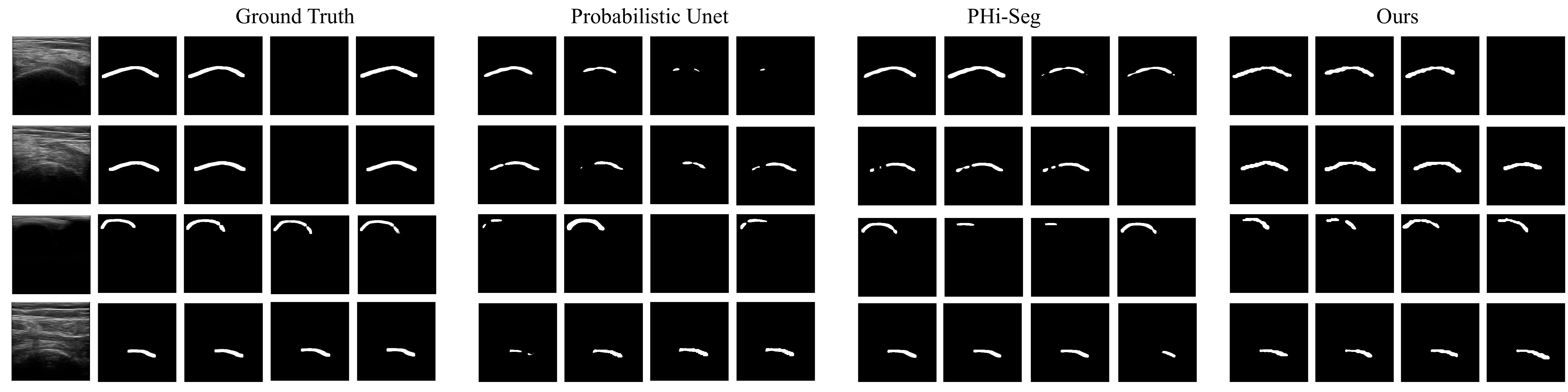}
\vspace{-2em}
 \caption{Comparative qualitative analysis with the two baseline methods Probabilistic U-net \cite{kohl2018probabilistic} and PHi-Seg \cite{baumgartner2019phiseg}. Examples of the Bone-US dataset with 1 expert and 3 novice gradings are shown on the left. The bone-US dataset has a comparatively small inter-rater disagreement. We sample the first 4 segmentation masks from prediction distribution.}
\label{Fig:bone}
\vskip -10 pt
\end{figure*}

\vskip-3pt
\noindent \textbf{Diversity Agreement.} For ambiguous models, the evaluation of the diversity of the predicted outputs can be tricky. Disproportionally rewarding diversity regardless of their match with the ground truth samples can be potentially misleading. On the other hand, the lack of diversity in predicted samples can indicate that model is rather deterministic than stochastic. Hence we consider matching maximum and minimum variance between two raters as they indicate minimum agreement and maximum agreement between two raters respectively. Here, we first calculate the variance between all pairs in ground truth distribution for a single input image. Then, we take the minimum and maximum variance. We define the minimum variance as  $V^{Y}_{min}$ and maximum variance as $V^{Y}_{max}$. Similarly, we calculate the variance between all pairs in the prediction distribution for that input and take the minimum and maximum variance. We define them as $V^{\hat{Y}}_{min}$ and $V^{\hat{Y}}_{max}$ respectively. The difference between the minimum variance of ground truth and prediction distribution for a particular input can be expressed as $\Delta V_{min} = |V^{Y}_{min} - V^{\hat{Y}}_{min}|$. Similarly, the difference between the maximum variance of ground truth and prediction distribution for a particular input is expressed as $\Delta V_{max} = |V^{Y}_{max} - V^{\hat{Y}}_{max}|$. Finally, we define the diversity agreement $D_a$ as,
\begin{equation}
      D_a = 1 - \left ( \frac{\Delta V_{max}+\Delta V_{min}}{2} \right).
\end{equation}

\begin{figure*}[htbp]

\centering
\includegraphics[width=1\linewidth]
{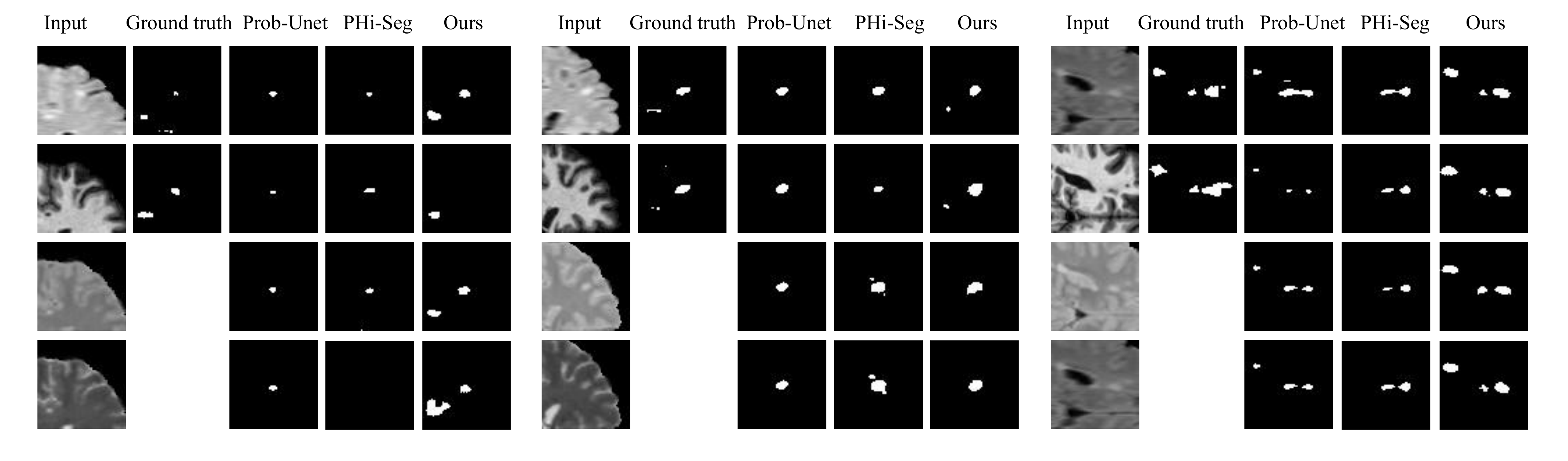}

\vspace{-1.5em}
 \caption{
Comparative qualitative analysis with the two baseline methods Probabilistic U-net \cite{kohl2018probabilistic} and PHi-Seg \cite{baumgartner2019phiseg}. Examples of the MS-MRI dataset with 2 expert gradings are shown here. We sample the first 4 segmentation masks from the prediction distribution.
}
\vskip -15pt
\label{Fig:mri}
\end{figure*}

\vspace{-2em}
\section{Experiments}

\subsection{Datasets}
\noindent \textbf{Lung lesion segmentation (LIDC-IDRI).} This publicly available dataset contains 1018 lung CT scans from 1010 subjects with manual annotations from four domain experts making it a good representation of typical CT image ambiguity \cite{armato2004lung}. A total of 12 radiologists provided annotation masks for this dataset. We use the dataset after the second reading where the experts were shown the annotations of other radiologists that allowed them to make new adjustments. The training set contains 13511 and the test set contains 1585 lesion images with 4 expert gradings.

\noindent \textbf{Bone surface segmentation (B-US).} Bone segmentation from ultrasound (US) imaging often results in a low inter-rater agreement \cite{hacihaliloglu2017ultrasound}. After obtaining institutional review board (IRB) approval 2000 US scans were collected from 30 healthy subjects. The scans were collected using a 2D C5-2/60 curvilinear probe and an L14-5 linear probe using the Sonix-Touch US machine (Analogic Corporation, Peabody, MA, USA). Depth settings and image resolutions varied between 3–8 cm, and 0.12–0.19 mm, respectively. All the collected scans were manually segmented by an expert ultrasonographer and three novice users who were trained to perform bone segmentation.  The training set contains 1769 and the test set contains 211 bone ultrasound scans. 
\noindent \textbf{Multiple sclerosis lesion segmentation (MS-MRI).} This publicly available dataset contains 84 longitudinal MRI scans from 5 subjects with a mean of 4.4 time-points \cite{carass2017longitudinal}. The white matter lesions associated with MS are delineated by two domain expert raters one with four and the other with ten years of experience. Both experts were blinded to the temporal ordering of the MRI scans. From the volumetric MRI, we convert each slice into a 2D image with corresponding segmentation masks. Each data point contains proton density (PD), Flair, MP RAGE, and T2 MRI scans. The training set contains 6012 from 4 patients and the test set contains 1411 scans from 1 patient.
\subsection{Implementation Details}
\vskip -6pt
The proposed method is implemented using the PyTorch framework. We set the time step as $T=1000$ with a linear noise schedule for all the experiments. The U-Net-like diffusion model’s weights and biases are optimized using an Adam optimizer with a learning rate of $10^{-4}$. For all the experiments $\lambda = 0.001$ is set to regularize $\mathcal{L}_{vlb}$. We also set $\beta = 0.001$ to prevent $\mathcal{L}_{amb}$ from overwhelming both $\mathcal{L}_{simple}$ and $\mathcal{L}_{vlb}$. We chose $128 \times 128$ as the resolution for LIDC-IDRI, $256 \times 256$ as the resolution for Bone-US, and $64 \times 64$ as the resolution for the MS-MRI dataset. For MS-MRI we concatenate all four MRI scans and use them as an image prior before feeding it to the diffusion model. The rest of the parameters for diffusion in the model are the same as \cite{nichol2021improved}.

\begin{table*}[]
	\centering
		\caption{
Comparison of quantitative results in terms of GED, CI, and $D_{max}$ for all the datasets with state-of-the-art ambiguous segmentation networks. The best results are in \textbf{Bold} and we achieve state-of-the-art results in terms of $D_{max}$ and CI score across all datasets.
  }
  \vskip -10pt
	\resizebox{1.9\columnwidth}{!}{
		\begin{tabular}{
				>{\columncolor[HTML]{FFFFFF}}c |
				>{\columncolor[HTML]{FFFFFF}}c 
				>{\columncolor[HTML]{FFFFFF}}c 
				>{\columncolor[HTML]{FFFFFF}}c |
				>{\columncolor[HTML]{FFFFFF}}c 
				>{\columncolor[HTML]{FFFFFF}}c 
				>{\columncolor[HTML]{FFFFFF}}c |
				>{\columncolor[HTML]{FFFFFF}}c 
				>{\columncolor[HTML]{FFFFFF}}c 
				>{\columncolor[HTML]{FFFFFF}}c }
			{\color[HTML]{000000} Method} & \multicolumn{3}{c|}{\cellcolor[HTML]{FFFFFF}{\color[HTML]{000000} LIDC-IDRI \cite{armato2004lung}}} & \multicolumn{3}{c|}{\cellcolor[HTML]{FFFFFF}{\color[HTML]{000000} Bone Segmentation }} & \multicolumn{3}{c}{\cellcolor[HTML]{FFFFFF}{\color[HTML]{000000} MS-Lesion \cite{carass2017longitudinal}}} \\ \hline
			& GED ($\downarrow$) & CI($\uparrow$)   & $D_{max}$($\uparrow$) & GED ($\downarrow$) & CI($\uparrow$) & $D_{max}$($\uparrow$) & GED ($\downarrow$) & CI($\uparrow$) & $D_{max}$($\uparrow$) \\ \hline
			Probabilistic Unet\cite{kohl2018probabilistic} & 0.353 & 0.731 & 0.892 & 0.390 & 0.738 & 0.844 & 0.749 & 0.514 & 0.502  \\ 
			{\color[HTML]{000000} PHi-Seg \cite{baumgartner2019phiseg}} & {\color[HTML]{000000} \textbf{0.270}} & {\color[HTML]{000000}  0.736} & {\color[HTML]{000000} 0.904} & {\color[HTML]{000000} 0.312} & {\color[HTML]{000000} 0.7544} & {\color[HTML]{000000} 0.848} & {\color[HTML]{000000} 0.681} & {\color[HTML]{000000} 0.518} & {\color[HTML]{000000} 0.506}  \\
			Generalized Probabilistic U-net  \cite{bhat2022generalized}& 0.299 & 0.707 & 0.905 & \textbf{0.289} & 0.7501 & 0.863 & \textbf{0.678} & 0.522 & 0.513  \\ 

			\hline
			
			{\color[HTML]{000000} $CIMD$ (Ours)} & {\color[HTML]{000000} 0.321} & {\color[HTML]{000000} \textbf{0.759}} & {\color[HTML]{000000} \textbf{0.915}} & {\color[HTML]{000000} 0.295} & {\color[HTML]{000000} \textbf{0.7578}} & {\color[HTML]{000000} \textbf{0.889}} & {\color[HTML]{000000} 0.733} & {\color[HTML]{000000} \textbf{0.560}} & {\color[HTML]{000000} \textbf{0.562}} 
		\end{tabular}
	}

	\label{quanres}
\vskip -15pt
\end{table*}
\vspace{-.5em}
\subsection{Evaluation Metrics}
\vskip -6 pt
\noindent \textbf{Generalized Energy Distance.} A commonly used metric in ambiguous image segmentation tasks that leverages distance between observations by comparing the distribution of segmentations \cite{kohl2018probabilistic}. It is given by \cite{bellemare2017cramer,salimans2018improving,szekely2013energy},
\begin{equation}
D^2_{GED}(P_{gt},P_{out}) = 2\mathbb{E}[d(S,Y)]-\mathbb{E}[d(S,S')]-\mathbb{E}[d(Y,Y')],
\end{equation}
where, $d$ corresponds to the distance measure $d(x, y) = 1 - IoU(x, y)$, $Y$ and $Y'$ are independent samples of $P_{gt}$ and $S$ and $S'$ are sampled from $P_{out}$. Lower energy indicates better agreement between prediction and the ground truth distribution of segmentations.

\vspace{-.7em}
\subsection{Comparison with the Baseline methods.}

To the best of our knowledge, there exists no other work that has considered explicitly modeling the ambiguity of medical images to produce multiple segmentation maps using diffusion models. We compare our approach with current state-of-the-art methods that are specifically designed to capture a distribution over multi-modal segmentation.

\noindent \textbf{Probabilistic U-net and its variants.} We report the results for the current state-of-the-art method for ambiguous medical image segmentation network probabilistic U-net \cite{kohl2018probabilistic}. We train a probabilistic U-net for the LIDC-IDRI dataset using the same parameter reported in the paper. For Bone-US and MS-MRI, we train the probabilistic U-net with $\beta = 10$ until the loss doesn’t improve. Additionally, we compare with a variant of probabilistic U-net, namely generalized probabilistic U-net \cite{bhat2022generalized}, where instead of using axis-aligned Gaussian distribution to model prior and posterior network, they use a mixture of full covariance Gaussian distributions.

\noindent \textbf{PHi-Seg.}  One of the major problems of probabilistic U-net is the lack of diversity in the predicted sample, as the stochasticity is only injected into the highest resolution. To solve this, PHi-Seg \cite{baumgartner2019phiseg} adopts a hierarchical structure inspired by Laplacian Pyramids, where the model generated conditional segmentation by refining the distribution at increasingly higher resolution, hence producing better quality samples. 
We train PHi-Seg using the parameters reported in the paper for all the datasets.

\noindent \textbf{Quantitative Comparison:} We consider both GED and CI scores to evaluate the performance of different ambiguous medical image segmentation models. We tabulate the results in Table \ref{quanres} evaluating 4 samples from the prediction distribution. We also separately report maximum Dice-matching scores. Our method outperforms other state-of-the-art networks in terms of both $D_{max}$ and CI scores. In terms of GED, we get on par performance with the Probabilistic U-net. A high $D_{max}$ score indicates the generated samples are in good match with the ground truth distribution and the CI score ensures the sample diversity matches with ground truth diversity.

\noindent \textbf{Qualitative Comparison:} As the evaluation of ambiguous networks is difficult to characterize, we argue that qualitative results can be a good indicator of network performance, especially for difficult cases. We show the predictions from the test dataset for all the models in Figure \ref{Fig:lidc}, \ref{Fig:bone}, and \ref{Fig:mri}. It can be seen that CIMD achieves visually superior and diverse results compared to the previous state-of-the-art methods. From Figure \ref{Fig:lidc} it can be observed that the model was able to capture the frequencies of blanks as well as maintained diversity in difficult cases. CIMD works especially well on ultrasound modalities with minimal error as can be seen in Figure \ref{Fig:bone}. From Figure \ref{Fig:mri} it can be seen that CIMD is able to capture all the lesions even if they have small structures while maintaining diversity in segmentation masks. As CIMD injects stochasticity at each hierarchical feature representation, it demonstrates diverse and accurate segmentation in all datasets.

\vspace{-1em}
\section{Discussion}

\begin{table}[]
	\centering
		\caption{Ablation study: We perform an ablation study on LIDC-IDRI dataset \cite{armato2004lung} to better understand the contributions incorporated in the CIMD method.}
		\vskip -5 pt
		\resizebox{0.75\linewidth}{!}{%
	\begin{tabular}{c|c|c|c}
		\Xhline{2\arrayrulewidth}

		Method           & GED ($\downarrow$)                       & CI ($\uparrow$) & $D_{max}$ ($\uparrow$)    \\ \Xhline{2\arrayrulewidth}

		DDPM-det-Seg \cite{wolleb2021diffusion} &  1.081    & 0.616     & 0.548                 \\
		 DDPM-Prob-Seg             & 0.417 & 0.683     & 0.689                 \\
		CIMD (Ours)            & \textbf{0.321}  & \textbf{0.759}     & \textbf{0.915}                 \\
		\Xhline{2\arrayrulewidth}

	\end{tabular}
	}
	\label{ablation1}
\vskip -15 pt
\end{table}

\vskip -5 pt
\noindent \textbf{Ablation Study} For the ablation study, we evaluate the proposed contribution. In particular, we compare our contribution with the original DDPM Model, which is inherently stochastic, hence capable of generating several segmentation masks from a single input image. We demonstrate the contribution in detail in our ablation study.

\noindent \textbf{DDPM for segmentation.} We report two sets of results for DDPM-based segmentation. \textit{DDPM-det-Seg} refers to the model where DDPM is trained using the average of all segmentation masks for an input image. \textit{DDPM-Prob-Seg} refers to training the DDPM with different segmentation masks for an input image. The results for the LIDC dataset can be observed in Table \ref{ablation1}. It can be observed that even though the DDPM sampling process is stochastic, the distribution of generated segmentation masks is not diverse enough as well as are not similar to the ground truth distribution. Additional ablation results are in supplementary.

\noindent \textbf{Limitations:} Although our proposed method produces diverse-yet-meaningful predictions, it is quite slow at training and inference due to the inherent nature of the diffusion process. Also, training a model for ambiguous segmentation requires annotations from multiple radiologists which is costly and time-consuming. In the future, we plan on extending our method trying to tackle these limitations while also extending the approach to more modalities and 3D volumetric datasets.
\vspace{-1em}
\section{Future work and Conclusion}
\vskip -8 pt
\noindent In this work, we introduce a diffusion-based ambiguous segmentation network that can generate multiple plausible annotations from a single input image. Unlike traditional cVAE-based networks, CIMD uses its hierarchical structure to incorporate stochasticity at each level and doesn’t require a separate network to encode prior information about the image during the inference stage. Our method can be incorporated into any diffusion-based framework with minimal additional training. For future work, it is possible that CIMD can be extended to more general computer vision problems as well as tested for other medical imaging modalities. Lastly, our approach can be incorporated in other specialized diffusion-based segmentation networks, e.g. MedSegDiff \cite{wu2022medsegdiff}, SegDiff\cite{wu2022medsegdiff}, etc for higher fidelity segmentation outputs.

{\small
\bibliographystyle{ieee_fullname}
\bibliography{egbib}
}

\clearpage
\pagebreak

\section*{A. Appendix Ablation Study}
\begin{figure*}[h!]
\centering
\includegraphics[width=1\linewidth]
{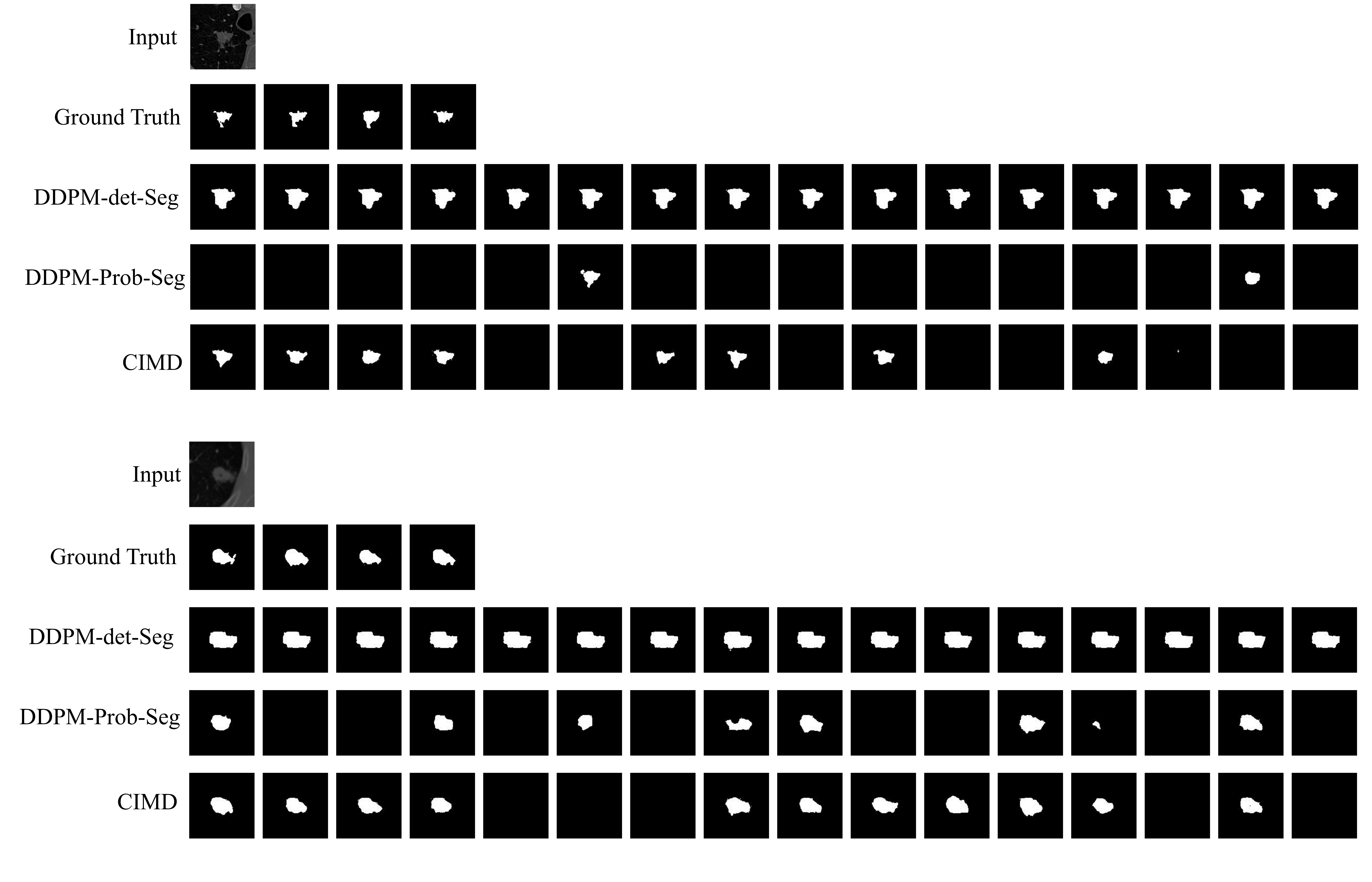}
 \caption{
Visualization of ablation study for LIDC-IDRI \cite{armato2004lung} dataset. DDPM-det-Sg is trained using the average of all segmentation masks of one input image. DDPM-Prob-Seg is trained using all segmentation masks of one input image.
}
\label{Fig:abl}
\end{figure*}

We additionally report the ablation study for both the Bone-US dataset and MS-MRI \cite{carass2017longitudinal} dataset. As we can observe from Table \ref{ablation2} and Table \ref{ablation3} that CIMD improves the diffusion model performance in terms of both GED and CI scores. We visualize the ablation study results for the LIDC-IDRI dataset \cite{armato2004lung} in Figure \ref{Fig:abl}. DDPM-det-Seg is the diffusion model \cite{nichol2021improved,wolleb2021diffusion} trained using the average of all four segmentation masks. Although the sampling process is stochastic, we see minimal changes in generated segmentation masks. DDPM-Prob-Seg is trained using all the segmentation masks. In other words, different segmentation masks are used in each forward pass for an input image. It can be seen that although there are some variations in segmentation masks, most of them are empty. In contrast to that, CIMD is able to segment the lesion as well as produce different segmentation masks that match the ground truth distributions. This proves DDPM itself is not able to model the stochasticity of the dataset alone.

\begin{table}[h!]
	\centering
		\caption{Ablation study: we perform an ablation study on the Bone-US dataset to better understand the contributions incorporated in the CIMD method.}
		\resizebox{0.8\linewidth}{!}{%
	\begin{tabular}{c|c|c|c}
		\Xhline{2\arrayrulewidth}

		Method           & GED ($\downarrow$)                       & CI ($\uparrow$) & $Dice_{max}$ ($\uparrow$)    \\ \Xhline{2\arrayrulewidth}

		DDPM-det-Seg \cite{wolleb2021diffusion} &   0.887   & 0.673     & 0.626                 \\
		 DDPM-Prob-Seg             & 0.798 & 0.675     & 0.627                 \\
		CIMD (Ours)            & \textbf{0.295}  & \textbf{0.757}     & \textbf{0.889}                 \\
		\Xhline{2\arrayrulewidth}

	\end{tabular}
	}
	\label{ablation2}

\end{table}

\begin{table}[h!]
	\centering
		\caption{Ablation study: We perform an ablation study on MS-MRI dataset \cite{carass2017longitudinal} to better understand the contributions incorporated in the CIMD method.}
		\resizebox{0.8\linewidth}{!}{%
	\begin{tabular}{c|c|c|c}
		\Xhline{2\arrayrulewidth}

		Method           & GED ($\downarrow$)                       & CI ($\uparrow$) & $Dice_{max}$ ($\uparrow$)    \\ \Xhline{2\arrayrulewidth}

		DDPM-det-Seg \cite{wolleb2021diffusion} &  0.799    & 0.507     & 0.497                 \\
		 DDPM-Prob-Seg             & 0.804 & 0.509     & 0.499                 \\
		CIMD (Ours)            & \textbf{0.733}  & \textbf{0.560}     & \textbf{0.562}                 \\
		\Xhline{2\arrayrulewidth}

	\end{tabular}
	}
	\label{ablation3}

\end{table}


\section*{B. Appendix Network Architecture}

\noindent \textbf{AMN and ACN architecture.} AMN (Ambiguity Modeling Network) and ACN (Ambiguity Controlling Network) have the same architecture, which is an encoder consisting of repeated application of four 3x3 convolution layers with 32, 64, 128, and 192 filters each followed by a rectified linear unit (ReLU) and a 2x2 average pooling with stride 2 for down-sampling. Then we add a 1x1 convolution layer that takes the global average pooled feature maps from the previous layer as input and predicts the Gaussian distribution which is parameterized by mean and variance. AMN takes the concatenation of the input image with the ground truths as input and predicts the Gaussian distribution of the segmentation masks conditioned on an input image. ACN takes the concatenation of the input image with the predictions as input and predicts the Gaussian distribution of predicted masks conditioned on the input image.

\section*{C. Appendix Training details}

\textbf{$\beta$ Parameter.} The regularization parameter $\beta$ is empirically chosen to be $0.001$, as higher $\beta$ overwhelms the other loss terms and produces noisy outputs. Lower $\beta$ that $0.001$ ignores the KL divergence between ACN and AMN, hence network acts like a regular diffusion model with minimal variations in outputs.

\subsection*{D. Appendix Qualitative Result Analysis}

\begin{figure*}[h!]

\centering
\includegraphics[width=1\linewidth]
{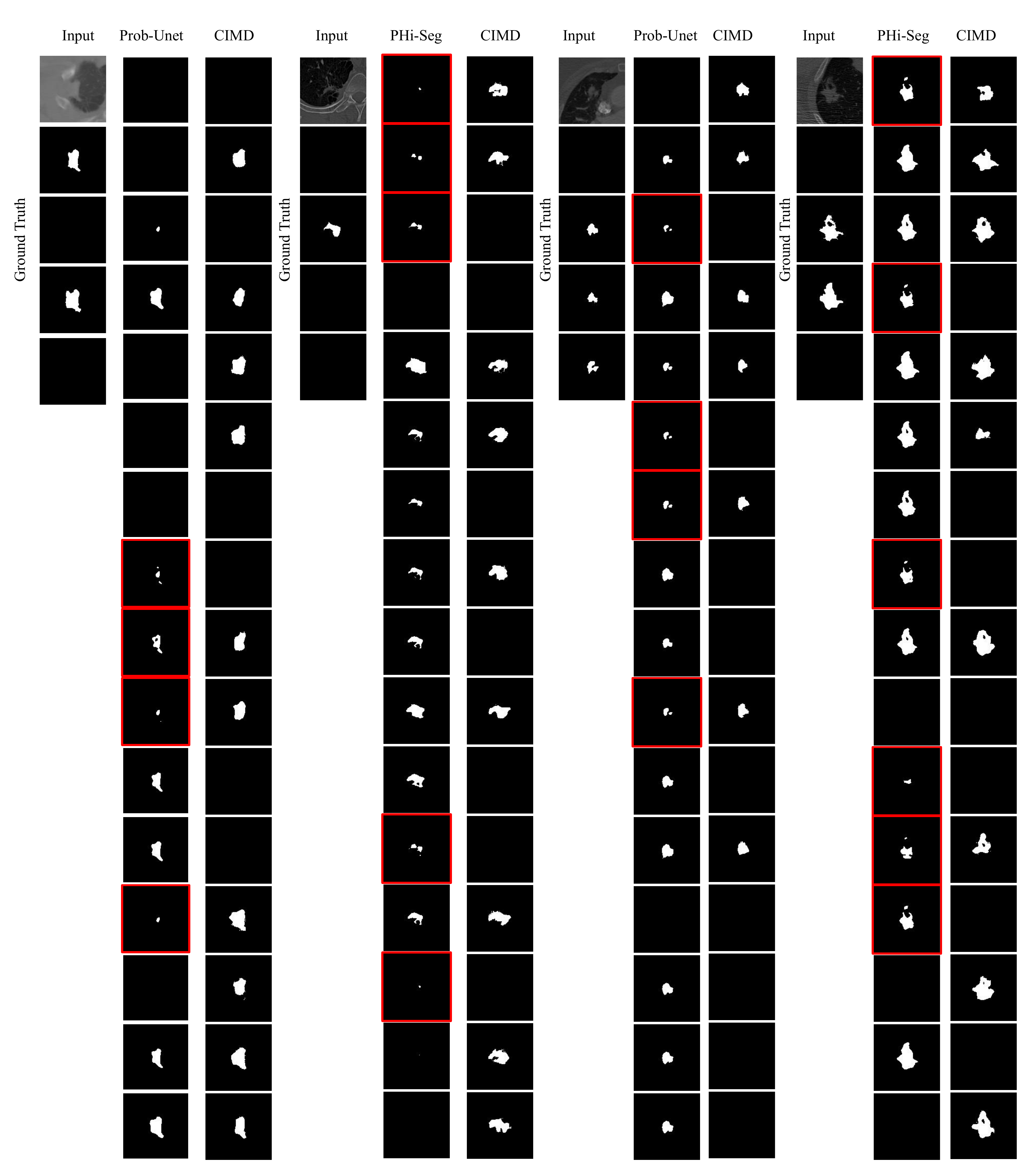}

 \caption{
Comparative qualitative analysis with the two baseline methods Probabilistic U-net \cite{kohl2018probabilistic} and PHi-Seg \cite{baumgartner2019phiseg} for LIDC-IDRI \cite{armato2004lung} dataset. Here we show 16 samples from each model. The \textcolor{red}{red} boxes indicate incomplete or noisy segmentation masks. Here we can observe some incomplete or noisy output from baseline methods while all 16 samples from CIMD have high fidelity.
}
\label{Fig:lidc_qual}
\end{figure*}

\begin{figure*}[htbp]

\centering
\includegraphics[width=1\linewidth]
{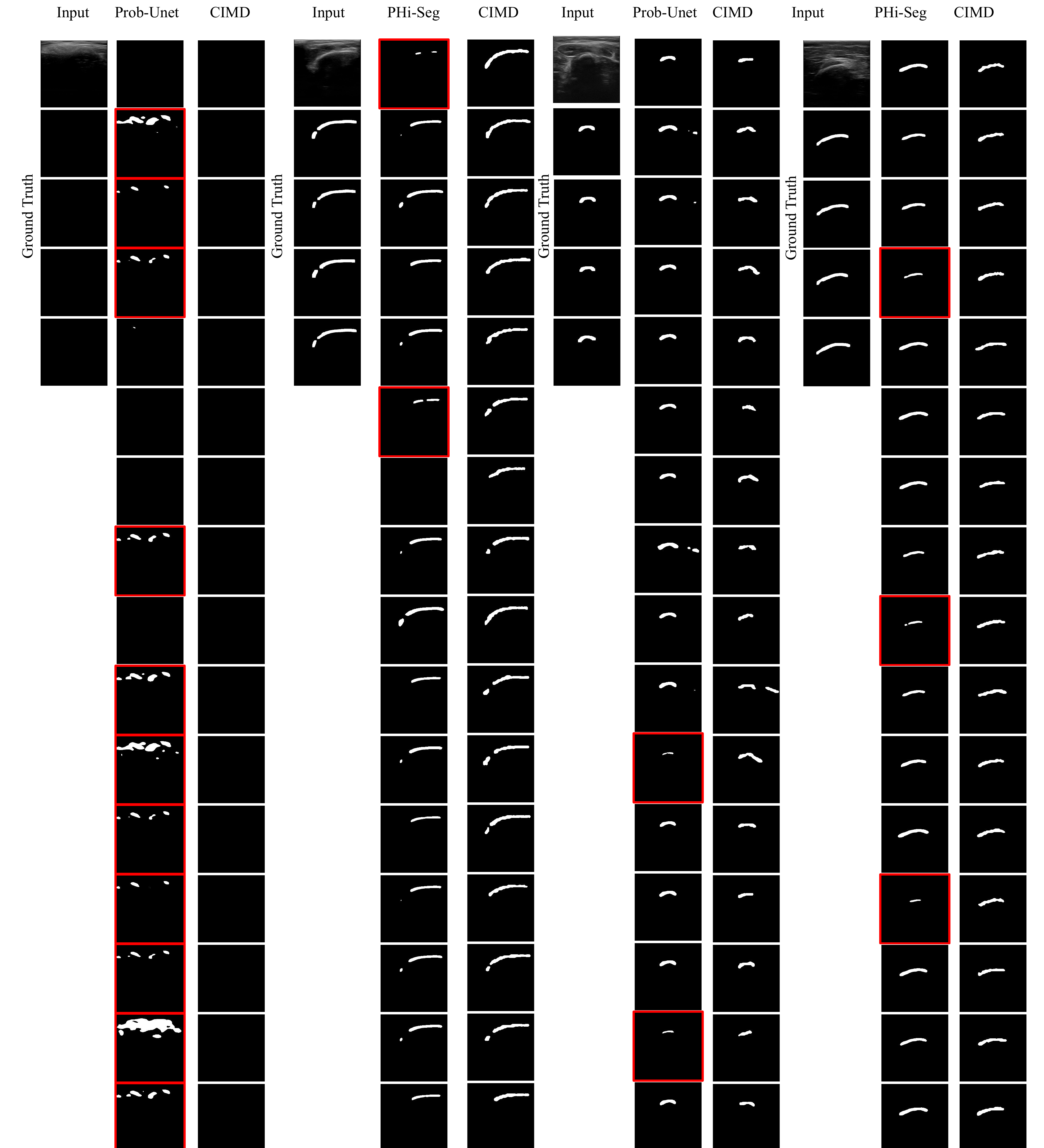}

 \caption{Comparative qualitative analysis with the two baseline methods Probabilistic U-net \cite{kohl2018probabilistic} and PHi-Seg \cite{baumgartner2019phiseg} for Bone-US dataset. Here we show 16 samples from each model. The \textcolor{red}{red} boxes indicate incomplete or noisy segmentation masks. Here we can observe some incomplete or noisy output from baseline methods while all 16 samples from CIMD have high fidelity.
}
\label{Fig:bone_qual}
\end{figure*}

\textbf{Average Segmentation Quality.} We visualize 16 samples for each input image from the test set distribution to assess their quality. For both Prob-Unet \cite{kohl2018probabilistic} and PHi-Seg \cite{baumgartner2019phiseg} we can observe from Figure \ref{Fig:lidc_qual} and Figure \ref{Fig:bone_qual} that although there are some segmentation masks that are close to ground truth (therefore, not affecting the quantitative metric much), not all segmentation masks are complete or consistent. This happens because they are sampled using different latent variables which might not always produce high-fidelity samples.
However, CIMD is observed to consistently produce high-fidelity samples as the model doesn't depend on latent variables from a prior model for segmentation.

\begin{figure*}[h!]
\centering
\includegraphics[width=1\linewidth]
{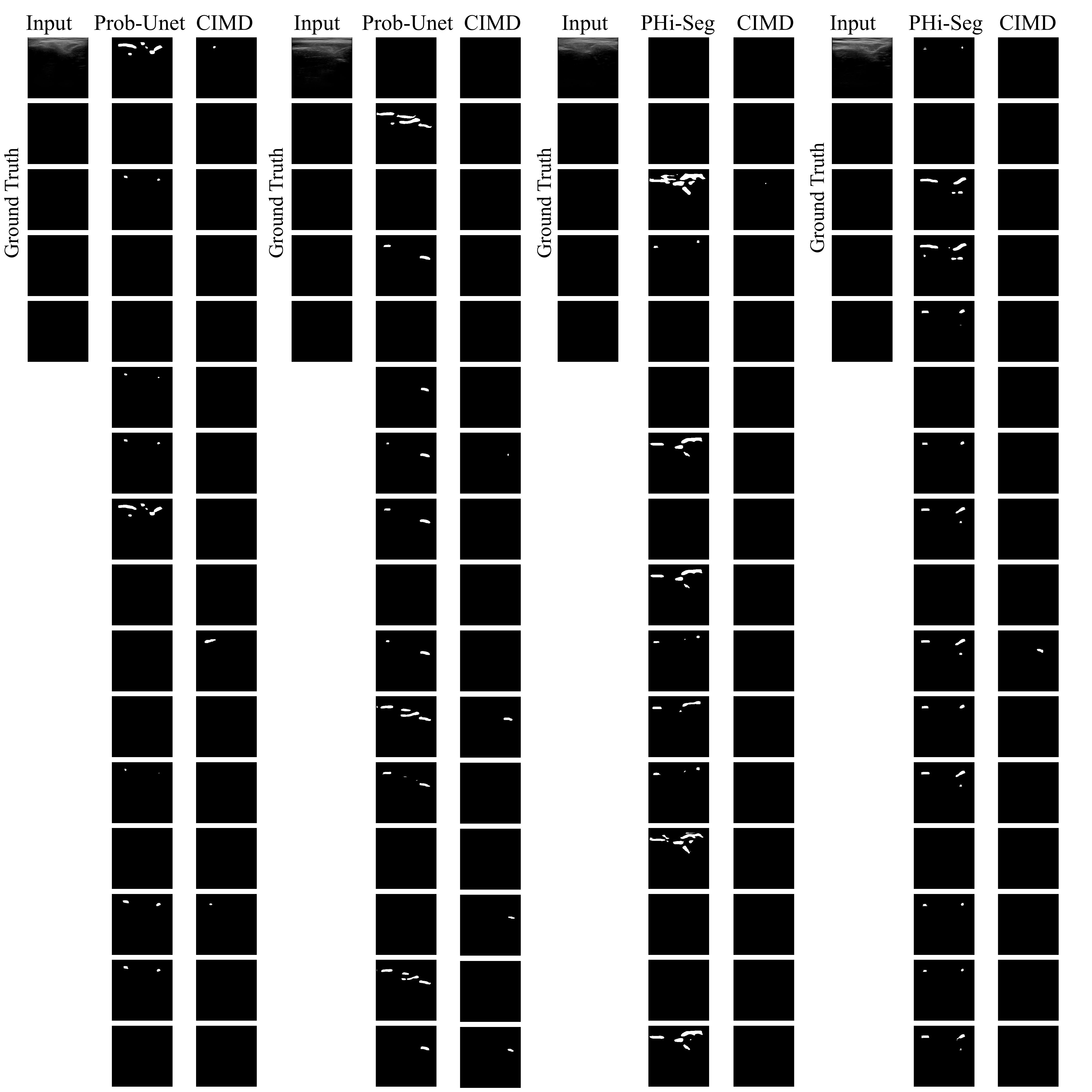}
 \caption{
Comparative qualitative analysis with the two baseline methods Probabilistic U-net \cite{kohl2018probabilistic} and PHi-Seg \cite{baumgartner2019phiseg} for blank segmentations from all experts in Bone-US dataset. We sample 16 masks from each model. We can observe that for blank annotation Prob-Unet and PHi-Seg both struggles as the noisy contrast resemble bone surface response in ultrasound images.
}
\label{Fig:blank_bone}
\end{figure*}

\textbf{Empty Segmentation in Bone-US dataset.} In ultrasound, the high acoustic impedance mismatch between soft tissue and bone surface produces a high contrast curve-linear region. This high-contrast region indicates the presence of the bone surface. However, this response can be extremely noisy due to the nature of ultrasound imaging. In our dataset, some ultrasound scan doesn't have any bone surface response, hence all four raters annotated them as empty masks. From Figure \ref{Fig:blank_bone} we can observe that some latent variables from both Prob-Unet and PHi-Seg struggle to ignore random contrast in ultrasound imaging, and segment those regions as bone surfaces. On the other hand, CIMD produces much more consistent results when the bone surface is not present with minimal error.

\begin{figure*}[h!]
\centering
\includegraphics[width=1\linewidth]
{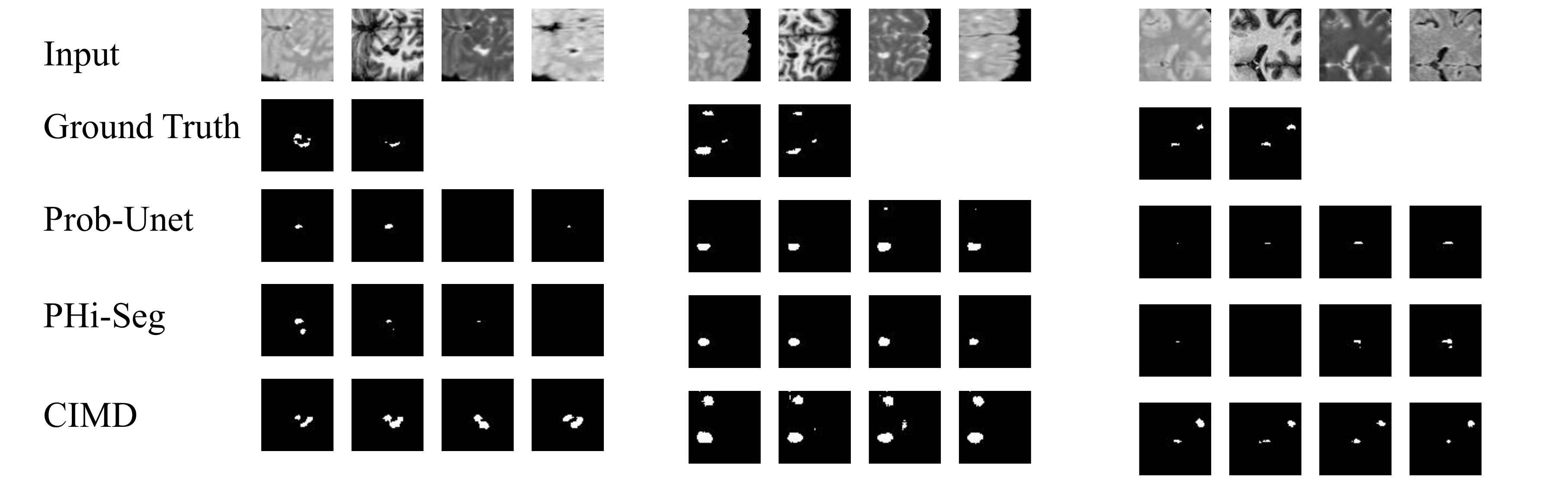}
 \caption{Qualitative comparison of MS-MRI \cite{carass2017longitudinal} dataset between Probabilistic U-net \cite{kohl2018probabilistic}, PHi-Seg \cite{baumgartner2019phiseg} and CIMD. We can observe that MS lesions have a very fine structure, hence both Prob-Unet and PHi-Seg are both failing to capture them. On the other hand, CIMD is able to capture even the smallest lesion that is present in the scan.
}
\label{Fig:mri_qual}
\end{figure*}

\textbf{Fine Lesion segmentation.} As MS-MRI \cite{carass2017longitudinal} dataset contains images with very fine lesions, it is difficult for other networks to segment it. However, from Figure \ref{Fig:mri_qual} it can be observed that CIMD is able to segment even the finest lesion from MRI scans.

\subsection*{E. Choice of Distribution}

\begin{table}[h!]
	\centering
		\caption{Quantitative results using LIDC-IDRI \cite{armato2004lung} dataset using CIMD with axis-aligned Gaussian (CIMD-AA) and full-covariance matrix (CIMD-FC).}
		\resizebox{0.8\linewidth}{!}{%
	\begin{tabular}{c|c|c|c}
		\Xhline{2\arrayrulewidth}

		Method           & GED ($\downarrow$)                       & CI ($\uparrow$) & $Dice_{max}$ ($\uparrow$)    \\ \Xhline{2\arrayrulewidth}

		CIMD-FC \cite{wolleb2021diffusion} &  0.447    & \textbf{0.774}     & 0.718                 \\

		CIMD-AA           & \textbf{0.321}  & 0.759     & \textbf{0.915}                 \\
		\Xhline{2\arrayrulewidth}

	\end{tabular}
	}
	\label{fc}

\end{table}

\begin{figure}[h!]
\centering
\includegraphics[width=1\linewidth]
{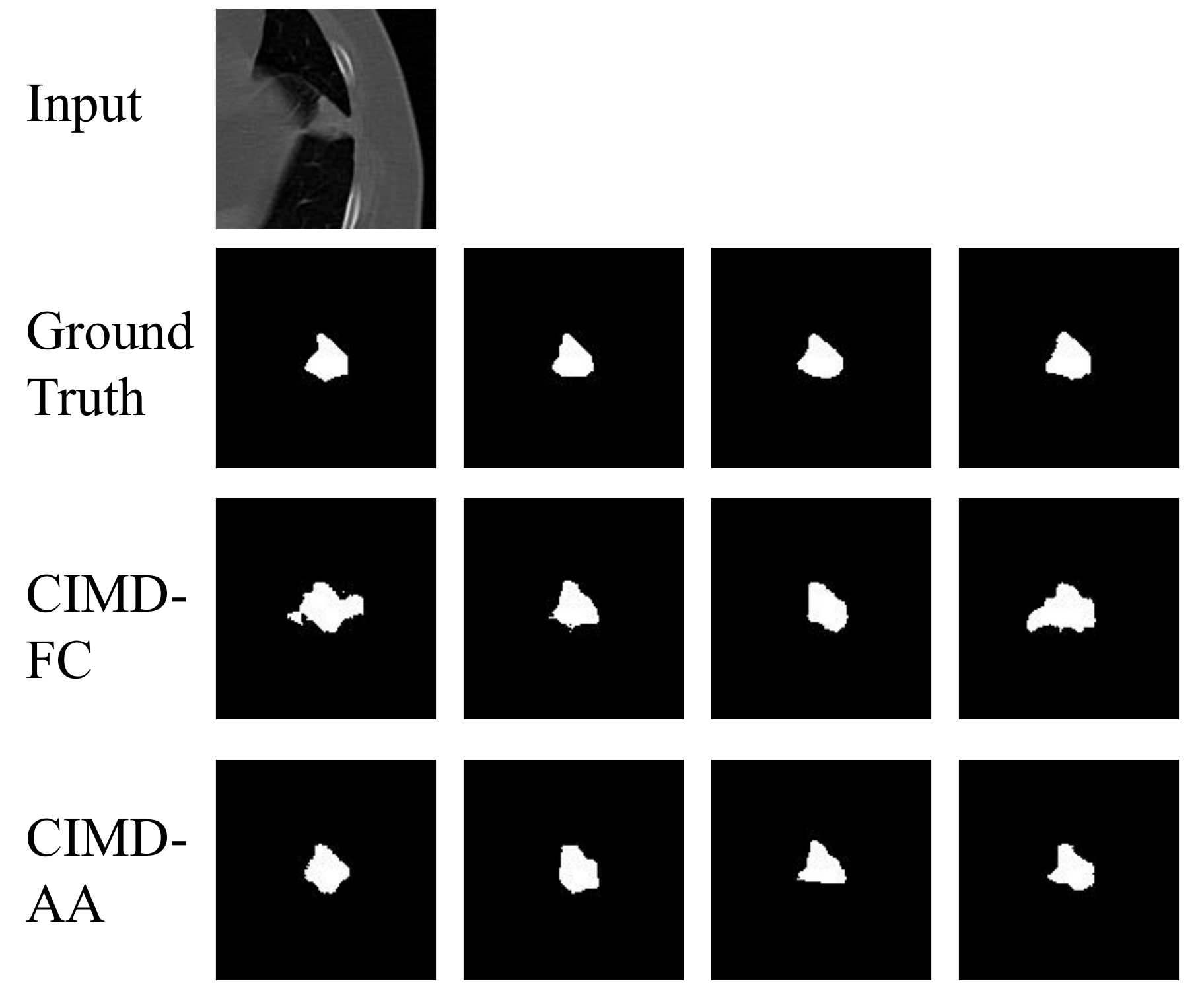}
 \caption{
Comparative qualitative analysis between CIMD-FC and CIMD-FC. CIMD-FC denotes CIMD with a full covariance matrix and CIMD-AA denotes CIMD with axis-aligned Gaussian.
}
\label{Fig:fc}
\end{figure}

In this section, we discuss the choice of distribution for AMN and ACN. The previous approach modeled the ambiguity of the segmentation masks using multivariate Gaussian with diagonal covariance matrix \cite{kohl2018probabilistic,baumgartner2019phiseg}. It has been assumed that the choice of a simple distribution restricts the sample diversity \cite{selvan2020uncertainty}. It has been hypothesized that the use of a full covariance matrix will produce a more diverse sample \cite{bhat2022generalized}. Generalized probabilistic U-net proposed the use of a full covariance matrix to model the distribution of segmentation masks \cite{bhat2022generalized}. Since the constraint of a valid covariance matrix is difficult to impose while training a network, the covariance matrix $\Sigma$ is built using Cholesky decomposition $L$ \cite{williams1996using}. 
\begin{equation}
    \Sigma = L L^T
\end{equation}
Here, $L$ is a positive valued diagonal lower-triangular matrix, which is computed by a neural network. The samples are drawn using the reparametrizing trick,

\begin{equation}
   z = \mu + L * \epsilon, \epsilon 	\sim \mathcal{N} (0, I) 
\end{equation}

Hence, we model the Gaussians of AMN and ACN with a full covariance matrix to observe its effect in the CIMD network. Although the CIMD with full covariance matrices were able to produce outputs with high diversity, they are always not close to ground truth distribution which can be observed from the $D_{max}$ in Table \ref{fc}. Moreover, CIMD with a full covariance matrix produces more coarse outputs hence the combined sensitivity is higher in this case. This skews the CI score, however, from $D_{max}$ and qualitative results in Figure \ref{Fig:fc} we can observe that axis-aligned performed better.

\end{document}